\pdfoutput=1
\documentclass{spie}
\usepackage{graphicx}
\usepackage{amsmath}
\usepackage{amssymb}
\usepackage{booktabs}
\usepackage{array}
\usepackage{xcolor}
\usepackage{float}
\usepackage{hyperref}
\usepackage{algorithm}
\usepackage{algpseudocode}

\title{A Semantics-First Approach to Multimodal Video Captioning with Frozen Encoders and Stabilized Reinforcement Learning}

\author{
    Albert Chiwara\supit{a} and Erhan Guven\supit{a}
    \skiplinehalf
    \supit{a}Johns Hopkins University, Baltimore, MD, USA
}

\authorinfo{Further author information: (Send correspondence to A. Chiwara)\\
A. Chiwara: E-mail: achiwar1@jh.edu\\
E. Guven: E-mail: eguven2@jhu.edu}

\begin{document}
\maketitle

{\NoHyper\let\thefootnote\relax\footnotetext{\textit{Note: This preprint corrects typographical errors regarding hyperparameter configurations and algorithmic notation, and clarifies baseline evaluation conditions present in the official conference proceedings. The empirical results and conclusions remain unchanged.}}\endNoHyper}

\begin{abstract}
We propose the Audio-Visual Query Transformer (AVQFormer), a highly efficient architecture built upon a 2.7B parameter LLM featuring a fixed-length $K=16$ semantic alignment module, stabilized via Adaptive Self-Critical Sequence Training (ASCST) with Entropy Regularization. Multimodal Large Language Models (LLMs) often rely on global audio representations to compensate for inefficient, frame-based visual encoders. While our Phase 1 prompted evaluations demonstrate that the Multimodal Base Model achieves strong deep semantic alignment (scoring 0.889 on BERTScore F1), optimizing this architecture for legacy metrics reveals a performance gap. We explore this gap as a hypothesis we term \textit{temporal alignment friction}: a conflict where global audio representations disrupt the chronological sequence of visual frames, leading to penalties under visually-focused evaluation metrics. Consequently, by stripping away the audio, our Video-Only ablation conforms more closely to this rigid syntax, achieving 63.0 CIDEr. While trailing larger models such as Qwen2.5-VL (67.4 CIDEr), this establishes a strong baseline for sub-3B parameter LLM-based models. This indicates that CIDEr, as applied to datasets such as MSR-VTT, may exhibit a bias toward visually grounded syntax, limiting its ability to fully capture multimodal semantic richness.
\end{abstract}


\section{Introduction}
As Multimodal Large Language Models scale, a common methodology is to fuse as many modalities as possible, typically injecting global audio embeddings to enrich the latent context provided by visual frames~\cite{donahue2015long, Yao2015}. However, standard n-gram metrics like CIDEr~\cite{Vedantam2015} exhibit a strong bias toward the literal, visually grounded syntax of human ground truths. In this manuscript, we empirically investigate how incorporating additional modalities affects literal captioning performance under standard metrics. We propose the AVQFormer, combining a strict $K=16$ temporal visual bottleneck with a lightweight 2.7B parameter LLM decoder (Phi-2~\cite{Javaheripi2023}). By successfully stabilizing Reinforcement Learning via Adaptive Self-Critical Sequence Training (ASCST)~\cite{Rennie2017} and Entropy Regularization~\cite{Holtzman2019}, we push the model to master both rigid dataset syntax and deep semantic meaning. Crucially, our experiments reveal that for highly optimized visual models, incorporating unaligned audio features introduces a performance gap under n-gram optimization objectives. While the audio pipeline retrieves factually accurate concepts, we hypothesize that it disrupts the precise chronological sequence of the visual storyboard, causing hallucinations or over-descriptions that legacy n-gram metrics punish. Consequently, we establish a highly competitive benchmark among lightweight ($\le$3B parameters) Video-LLMs on the MSR-VTT dataset~\cite{Xu2016, Chen2022CleanMSRVTT}. Our contributions are threefold: (1) a lightweight multimodal architecture with a strict temporal bottleneck, (2) a stabilized reinforcement learning framework (ASCST), and (3) an exploratory analysis of the performance gap between unimodal and multimodal models under n-gram optimization objectives.

\section{Related Work}
The field of video captioning has evolved rapidly, driven by advances in deep learning. This progression can be understood as a convergence of four key research threads: attentive sequence-to-sequence models, large-scale pretraining, modular large multimodal models (LMMs) with efficient adaptation, and reinforcement learning-based optimization.

\subsection{Foundational Paradigms: From Sequence-to-Sequence to Attention}
The architectural blueprint for modern captioning systems was established by the sequence-to-sequence (seq2seq) framework~\cite{sutskever2014sequence}. This approach was quickly adapted for visual tasks, with pioneering models using a Convolutional Neural Network (CNN) encoder and a Recurrent Neural Network (RNN) decoder~\cite{donahue2015long, vinyals2015show}. The information bottleneck inherent in fixed-length context vectors was addressed by the introduction of the attention mechanism~\cite{bahdanau2015neural}. ``Show, Attend and Tell''~\cite{xu2015show} applied this to image captioning, and it was swiftly extended to the temporal domain for video~\cite{Yao2015}.

\subsection{The Transformer Revolution and Large-Scale Pre-training}
The Transformer architecture, with its self-attention mechanism, superseded RNNs due to its superior ability to model long-range dependencies~\cite{vaswani2017attention}. Video-specific Transformer models, such as TimeSformer~\cite{Bertasius2021}, established pure self-attention-based architectures for video representation learning. A more significant development was the paradigm shift toward Vision-Language Pre-training (VLP), which fundamentally changed the goal of training from creating specialized models to building general-purpose foundation models. Contrastive learning on massive datasets of image-text pairs, exemplified by CLIP~\cite{Radford2021} and ALIGN~\cite{jia2021scaling}, yielded powerful, general-purpose visual representations. This paradigm has also been successfully applied to the audio domain, with models like CLAP~\cite{Wu2023} learning joint audio-text embeddings.

\subsection{Large Multimodal Models and Efficient Adaptation}
The most recent evolution involves connecting frozen foundation encoders with the generative capabilities of large language models (LLMs). A common architectural pattern in LMMs is the use of a trainable ``bridge'' module. A highly effective approach is the Query Transformer (Q-Former), introduced in BLIP-2~\cite{Li2023BLIP2}, which effectively aligns high-dimensional visual input with the textual embedding space using a fixed set of latent tokens. For video, this involves adapting the architecture to handle temporal dynamics, as seen in modern conversational agents. A key enabler of this modular paradigm is the development of Parameter-Efficient Fine-Tuning (PEFT) techniques, such as low-rank adaptation (LoRA)~\cite{Hu2021} and rank-stabilized LoRA (rsLoRA)~\cite{Kalajdzievski2023}, making the ``frozen encoder, frozen decoder'' architecture highly practical.

\subsection{Direct Metric Optimization via Reinforcement Learning}
Standard training using cross-entropy loss suffers from exposure bias and a metric mismatch. To address these issues, Reinforcement Learning (RL) provides a framework for directly optimizing non-differentiable sequence-level reward functions such as CIDEr~\cite{Vedantam2015}. Policy gradient methods, based on the REINFORCE algorithm~\cite{Williams1992}, maximize the final reward. A significant improvement was achieved with Self-Critical Sequence Training (SCST)~\cite{Rennie2017}, which uses the reward of the model's own greedy decoding as a baseline. This work builds on this foundation by developing a heavily stabilized version of SCST to improve training dynamics for massive generative LLMs.

\section{Methodology}

\subsection{Dense Semantic Alignment via AVQFormer}
To map high-dimensional spatio-temporal video data into the textual domain, we employ a Q-Former-inspired bottleneck~\cite{Li2023BLIP2}.

\begin{figure}[ht]
\centering
\includegraphics[width=0.9\linewidth]{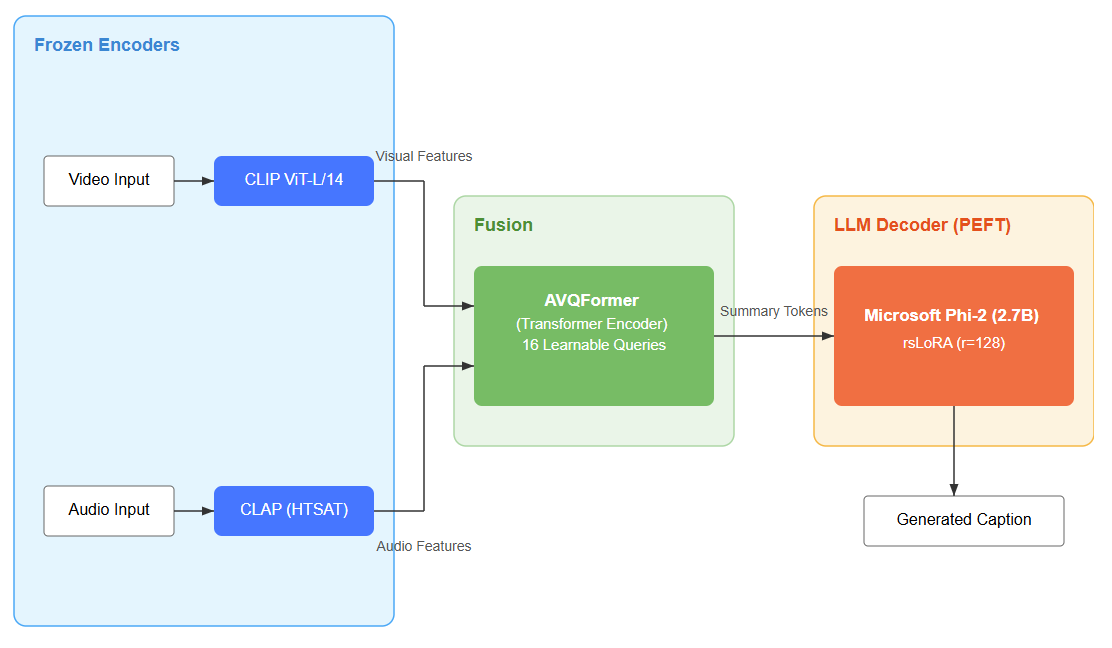}
\caption{Overall model architecture. The system comprises three main stages: Frozen Encoders for perception, a Trainable Fusion module for cross-modal alignment, and a parameter-efficiently tuned LLM Decoder for generation. \textit{(Note: AVQFormer acts as the Transformer Encoder, and Phi-2 serves as the generative engine.)}}
\label{fig:model-diagram}
\end{figure}

We extract frame-wise spatial features using a frozen OpenAI CLIP ViT-L/14 model. Specifically, we uniformly extract $T=16$ frames per video clip. For multimodal configurations, auditory features are extracted using a frozen HTSAT-based CLAP model, producing a single global acoustic anchor token. Let $\mathbf{F}_v \in \mathbb{R}^{T \times D_v}$ be the sequence of $T$ visual features extracted by CLIP, and $\mathbf{F}_a \in \mathbb{R}^{D_a}$ be the clip-level audio feature vector from CLAP. The features are first projected into a common embedding dimension $D_e$, compatible with the LLM:
\begin{align}
    \mathbf{P}_v &= \mathrm{Linear}_v(\mathbf{F}_v) \in \mathbb{R}^{T \times D_e}, \\
    \mathbf{P}_a &= \mathrm{Linear}_a(\mathbf{F}_a) \in \mathbb{R}^{1 \times D_e}.
\end{align}

Rather than temporally compressing the video, we utilize exactly $K=16$ learnable queries $\mathbf{Q} \in \mathbb{R}^{K \times D_e}$ to maintain a strict 1:1 temporal mapping with the extracted frames. This fixed-length configuration was empirically determined to be the optimal sweet spot for semantic distillation: extracting the most salient, LLM-aligned concepts from the raw visual features without disrupting the chronological storyboard. The queries, visual frames, and single audio token are concatenated directly:
\begin{equation}
    \mathbf{S}_{\text{in}} = \mathrm{Concat}(\mathbf{Q}, \mathbf{P}_v, \mathbf{P}_a) \in \mathbb{R}^{(K+T+1) \times D_e}.
\end{equation}

Through the encoder's self-attention mechanism, the queries interact with and align the most salient information from the multimodal audio-visual stream. The hidden states corresponding only to the original query positions are extracted from the encoder output $\mathbf{S}_{\text{out}}$:
\begin{equation}
    F_{\text{fused}} = \mathbf{S}_{\text{out}}[:K, :] \in \mathbb{R}^{K \times D_e}.
\end{equation}
This fixed-length sequence serves as an expressive, information-rich fused embedding that conditions the decoder.

\subsection{Adaptive SCST and Entropy Regularization}
Standard Teacher Forcing loops~\cite{sutskever2014sequence} produce models that are grammatically coherent but highly verbose and generic. We transition to Reinforcement Learning utilizing a modified Self-Critical Sequence Training (SCST)~\cite{Rennie2017} objective:

\begin{equation}
\nabla_\theta \mathcal{L}_{\text{policy}}(\theta) = - \mathbb{E}_{y^s \sim p_\theta} \left[ (r(y^s) - b_{\text{adapt}}) \nabla_\theta \log p_\theta(y^s) \right].
\end{equation}

To prevent the rapid mode collapse inherent in RL~\cite{Williams1992}, we introduce Adaptive SCST (ASCST), which dynamically clamps the baseline reward. The adaptive baseline $b_{\text{adapt}}$ is scaled by the ratio $\rho$ of the moving averages of past sampled rewards ($H^s$) to greedy rewards ($H^g$):

\begin{equation}
\rho = \frac{\mu(H^s)}{\mu(H^g) + \epsilon} \quad \implies \quad b_{\text{adapt}} = \beta \cdot \rho \cdot r(y^g),
\end{equation}

where $\beta$ is a damping factor and $r(y^g)$ is the greedy reward. Unlike standard SCST, which uses a fixed greedy baseline, ASCST dynamically rescales the baseline using reward statistics, improving stability under high-variance LLM decoding. In practice, to standardize the magnitude of the gradient updates and prevent destabilization, the raw advantage $A = r(y^s) - b_{\text{adapt}}$ is Z-score normalized across the batch dimension ($\hat{A}$) prior to computing the policy loss. Additionally, to penalize overly confident, repetitive policies, we apply an Entropy Regularization penalty~\cite{Holtzman2019}:

\begin{equation}
\mathcal{L}_{\text{entropy}} = -\lambda_E \left[ \frac{1}{B} \sum_{i=1}^B \left( \frac{\sum_{t=1}^L \mathcal{H}_{i,t} \cdot M_{i,t}}{\sum_{t=1}^L M_{i,t}} \right) \right],
\end{equation}

where $\mathcal{H}_{i,t}$ is the Shannon entropy of the token distribution for sequence $i$ at step $t$, $B$ is the batch size, $M_{i,t} \in \{0,1\}$ is the padding mask, and $\lambda_E$ is the regularization weight. The complete stabilized curriculum is formalized in Algorithm~\ref{alg:stabilized_scst}.

\begin{algorithm}[htbp]
\caption{Stabilized Self-Critical Sequence Training (SCST)}
\label{alg:stabilized_scst}
\begin{algorithmic}[1]
\Require Dataset $\mathcal{D}$, Batch size $B$, Learning rate $\eta$, Entropy weight $\lambda_E = 10^{-3}$, Baseline scalar $\beta = 0.9$, Window size $W = 200$, Small constant $\epsilon = 10^{-8}$
\State Initialize moving average queues $H^s \leftarrow \emptyset$, $H^g \leftarrow \emptyset$ of max length $W$
\For{each batch of videos $V=\{v_1, \dots, v_B\} \in \mathcal{D}$}
    \State \textbf{1. Rollout \& Reward Calculation}
    \State Sample captions $Y^s = \{y_1^s, \dots, y_B^s\}$ via Top-$p$ sampling ($p=0.9$)
    \State Generate greedy captions $Y^g = \{y_1^g, \dots, y_B^g\}$
    \State Calculate CIDEr rewards $R^s = \{r(y_1^s), \dots, r(y_B^s)\}$ and $R^g = \{r(y_1^g), \dots, r(y_B^g)\}$
    \State \textbf{2. Adaptive Baseline \& Advantage (ASCST)}
    \State Append batch means $\mu(R^s)$ and $\mu(R^g)$ to histories $H^s$ and $H^g$ respectively
    \State Calculate scale ratio: $\rho = \frac{\mu(H^s)}{\mu(H^g) + \epsilon}$
    \For{$i = 1 \dots B$}
        \State Compute adaptive baseline: $b_{\text{adapt}}^{(i)} = \beta \cdot \rho \cdot r(y_i^g)$
        \State Compute raw advantage: $A_i = r(y_i^s) - b_{\text{adapt}}^{(i)}$
    \EndFor
    \State \textbf{3. Per-Batch Z-Score Normalization}
    \State $\mu_A \leftarrow \frac{1}{B}\sum_{i=1}^B A_i$, \quad $\sigma_A \leftarrow \sqrt{\frac{1}{B}\sum_{i=1}^B (A_i - \mu_A)^2}$
    \State Normalize advantage: $\hat{A}_i = \frac{A_i - \mu_A}{\sigma_A + \epsilon}$ \quad for $i = 1 \dots B$
    \State \textbf{4. Token-Level Masking \& Policy Loss}
    \State Let $M_{i,t} \in \{0, 1\}$ be the padding mask for token $t$ in sequence $y_i^s$
    \State $\mathcal{L}_{\text{policy}} = - \frac{1}{B} \sum_{i=1}^B \left( \hat{A}_i \sum_t \log p_\theta(y_{i,t}^s \mid y_{i,<t}^s) \cdot M_{i,t} \right)$
    \State \textbf{5. Entropy Regularization \& Parameter Update}
    \State Let $\mathcal{H}_{i,t} = - \sum_{v \in \mathcal{V}} p_\theta(v \mid y_{i,<t}^s) \log p_\theta(v \mid y_{i,<t}^s)$
    \State $\mathcal{L}_{\text{entropy}} = - \lambda_E \left[ \frac{1}{B} \sum_{i=1}^B \left( \frac{\sum_t \mathcal{H}_{i,t} \cdot M_{i,t}}{\sum_t M_{i,t}} \right) \right]$
    \State Compute total loss: $\mathcal{L} = \mathcal{L}_{\text{policy}} + \mathcal{L}_{\text{entropy}}$
    \State Update AVQFormer and rsLoRA parameters: $\theta \leftarrow \theta - \eta \nabla_\theta \mathcal{L}$
\EndFor
\end{algorithmic}
\end{algorithm}

\section{Experiments and Results}

\subsection{Implementation Details}
To ensure reproducibility, we implemented the AVQFormer utilizing a strict $K=16$ sequence length for the semantic alignment module. All models were trained on a single NVIDIA RTX 5090 GPU. During Phase 1 Cross-Entropy training, this was achieved via a per-device batch size of 16 with 4 gradient accumulation steps (yielding an effective batch size of 64), whereas the Phase 2 SCST optimization utilized a per-device batch size of 16 with 2 gradient accumulation steps (yielding an effective batch size of 32). Parameter-efficient fine-tuning was conducted using rsLoRA~\cite{Kalajdzievski2023}. Due to the inherent regularization provided by the dataset's diverse multi-reference annotations, explicit dropout was omitted ($p=0.0$) across the AVQFormer and rsLoRA modules, with the architecture relying instead on Entropy Regularization during Phase 2 to prevent policy overfitting. To stabilize the Reinforcement Learning phase, we applied a 3-epoch SCST curriculum.

\subsection{State-of-the-Art on the MSR-VTT Test Set}
\begin{figure}[ht]
\centering
\includegraphics[width=0.8\textwidth]{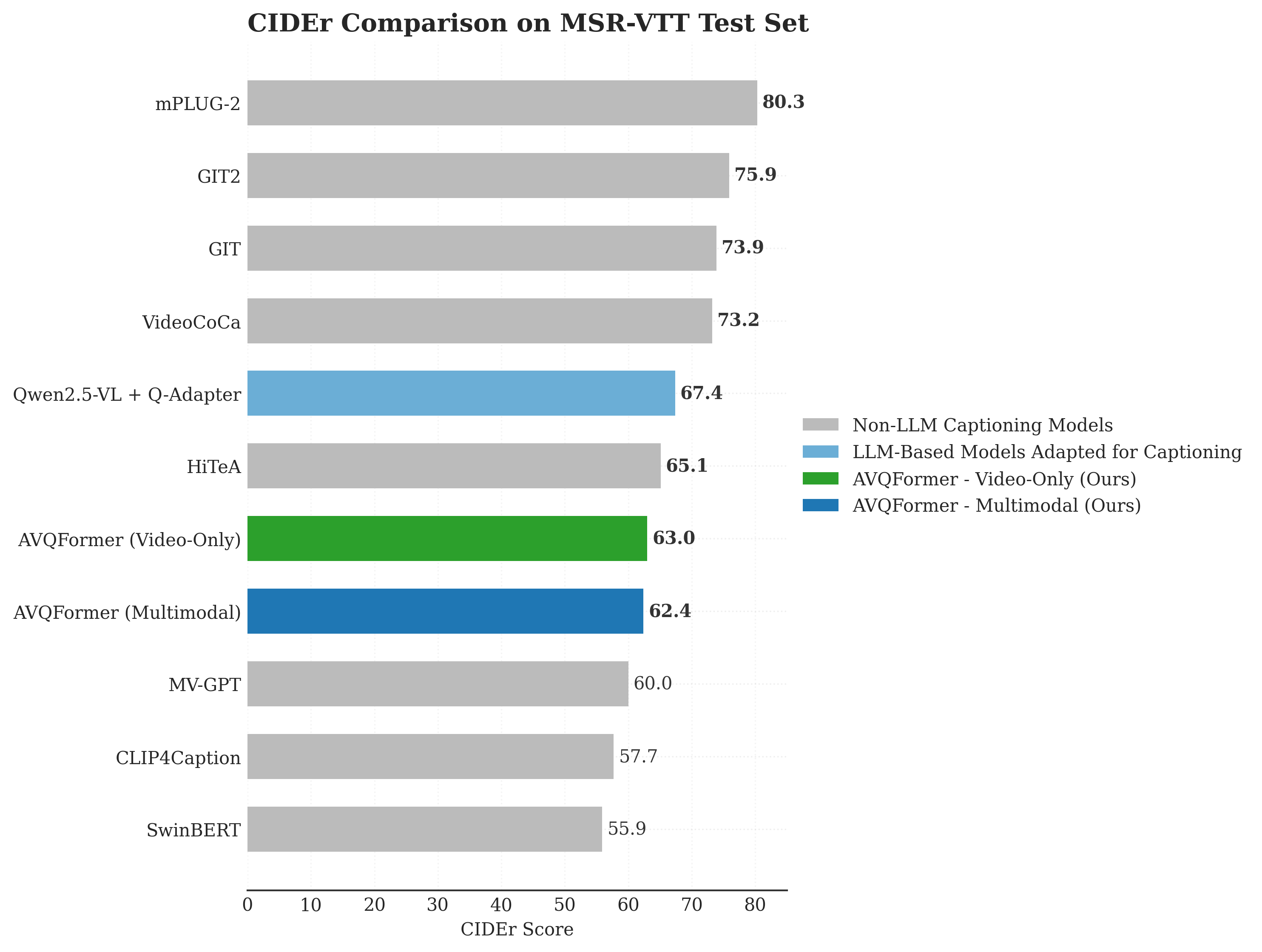}
\caption{CIDEr Score on MSR-VTT Test Set highlighting the structural divide between classical non-LLM captioning models and modern LLM-Based Models (Adapted for Captioning).}
\label{fig:sota_bar}
\end{figure}

We evaluate the proposed AVQFormer architectures on the MSR-VTT test split to enable direct comparison with prior work under standardized captioning metrics. All comparisons are restricted to models reporting MSR-VTT captioning metrics under comparable evaluation settings. As summarized in Table~\ref{tab:sota_msrvtt} and illustrated in Figure~\ref{fig:sota_bar}, the literature exhibits a clear structural divide: state-of-the-art CIDEr performance is dominated by classical non-LLM architectures (e.g., mPLUG-2, GIT2) that are explicitly optimized for n-gram overlap through supervised fine-tuning or reinforcement learning. In contrast, LLM-based multimodal models are less commonly evaluated under this strict syntactic protocol, and are often designed for zero-shot or instruction-following tasks rather than direct metric optimization.

Within this context, the AVQFormer achieves strong performance among lightweight ($\le$3B parameter) LLM-based models. The Video-Only variant reaches 63.0 CIDEr and 47.5 BLEU-4, demonstrating that raw base LLM architectures can remain competitive with mid-tier classical captioning models when explicitly optimized for captioning metrics via stabilized reinforcement learning. At the same time, among models reporting semantic evaluation metrics, AVQFormer variants attain BERTScore F1 scores of 0.9365 (Multimodal) and 0.9367 (Video-Only)~\cite{zhang2019bertscore}, indicating consistently high semantic alignment despite the constraints imposed by n-gram-based optimization. Notably, zero-shot baselines such as Panda-70M exhibit substantially lower CIDEr performance due to the absence of direct metric optimization.

Importantly, these results must be interpreted in light of differing training objectives. Models optimized with SCST (such as AVQFormer) directly maximize CIDEr and therefore benefit from alignment with the evaluation metric, whereas zero-shot or lightly adapted LLM baselines are not trained under this objective. Consequently, the comparison highlights not only performance differences, but also the fundamental tension between syntactic metric optimization and broader semantic modeling in multimodal LLMs. This reflects a fundamental metric-model mismatch: models optimized for semantic richness are evaluated using syntactic overlap metrics.

\begin{table*}[t]
\centering
\small
\caption{
MSR-VTT test set captioning performance comparing classical non-LLM architectures and LLM-Based Models (Adapted for Captioning).
Metrics are abbreviated as follows: \textbf{C} (CIDEr), \textbf{B@4} (BLEU-4), \textbf{M} (METEOR), \textbf{R} (ROUGE-L), and \textbf{BS} (BERTScore F1). All values are taken directly from the respective original papers.
Models are grouped by training paradigm: (i) classical captioning models optimized directly for n-gram metrics via supervised fine-tuning, and (ii) LLM-based multimodal models, including parameter-efficiently fine-tuned (PEFT) systems adapted for captioning and zero-shot baselines.
All scores correspond to the MSR-VTT test split unless otherwise specified.
This comparison highlights the structural gap between models explicitly optimized for syntactic metrics and modern LLM-based architectures, which are less commonly evaluated under such protocols.
}
\label{tab:sota_msrvtt}
\begin{tabular}{l l c c c c c}
\toprule
\textbf{Model} & \textbf{Setting} & \textbf{C} & \textbf{B@4} & \textbf{M} & \textbf{R} & \textbf{BS} \\
\midrule
\multicolumn{7}{c}{\textit{Classical Captioning Models (Non-LLM)}} \\
\midrule
mPLUG-2~\cite{Xu2023} & Fine-tuned & 80.30 & 57.80 & 34.90 & 70.10 & -- \\
GIT2~\cite{Wang2022GIT} & Fine-tuned & 75.90 & 54.80 & 33.10 & -- & -- \\
GIT~\cite{Wang2022GIT} & Fine-tuned & 73.90 & 53.80 & 32.90 & -- & -- \\
VideoCoCa~\cite{Yan2022VideoCoCa} & Fine-tuned & 73.20 & 53.80 & -- & 68.00 & -- \\
HiTeA~\cite{Ye2023HiTeA} & Fine-tuned & 65.10 & -- & 30.70 & -- & -- \\
MV-GPT~\cite{Seo2022MVGPT} & Fine-tuned & 60.00 & 48.92 & 38.66 & 64.00 & -- \\
CLIP4Caption~\cite{Tang2021} & Fine-tuned & 57.70 & 46.10 & 30.70 & 63.70 & -- \\
SwinBERT~\cite{Lin2022SwinBERT} & Fine-tuned & 55.90 & -- & 29.90 & -- & -- \\
\midrule
\multicolumn{7}{c}{\textit{LLM-Based Models (Adapted for Captioning)}} \\
\midrule
Qwen2.5-VL + Q-Adapter~\cite{Qwen25VL2025} & PEFT (adapted for captioning) & 67.38 & 49.84 & 32.32 & 62.69 & -- \\
Panda-70M (Student)~\cite{Panda70M} & Zero-shot & 31.50 & 25.40 & 27.70 & 50.10 & 0.8790 \\
AVQFormer (Multimodal) & PEFT (2.7B LLM) & 62.39 & 46.44 & 30.63 & 64.17 & 0.9365 \\
AVQFormer (Video-Only) & PEFT (2.7B LLM) & 63.00 & 47.50 & 30.77 & 64.47 & 0.9367 \\
\bottomrule
\end{tabular}
\end{table*}

\subsection{Ablation Studies (Validation Set)}
\begin{figure}[ht]
\centering
\includegraphics[width=0.7\textwidth]{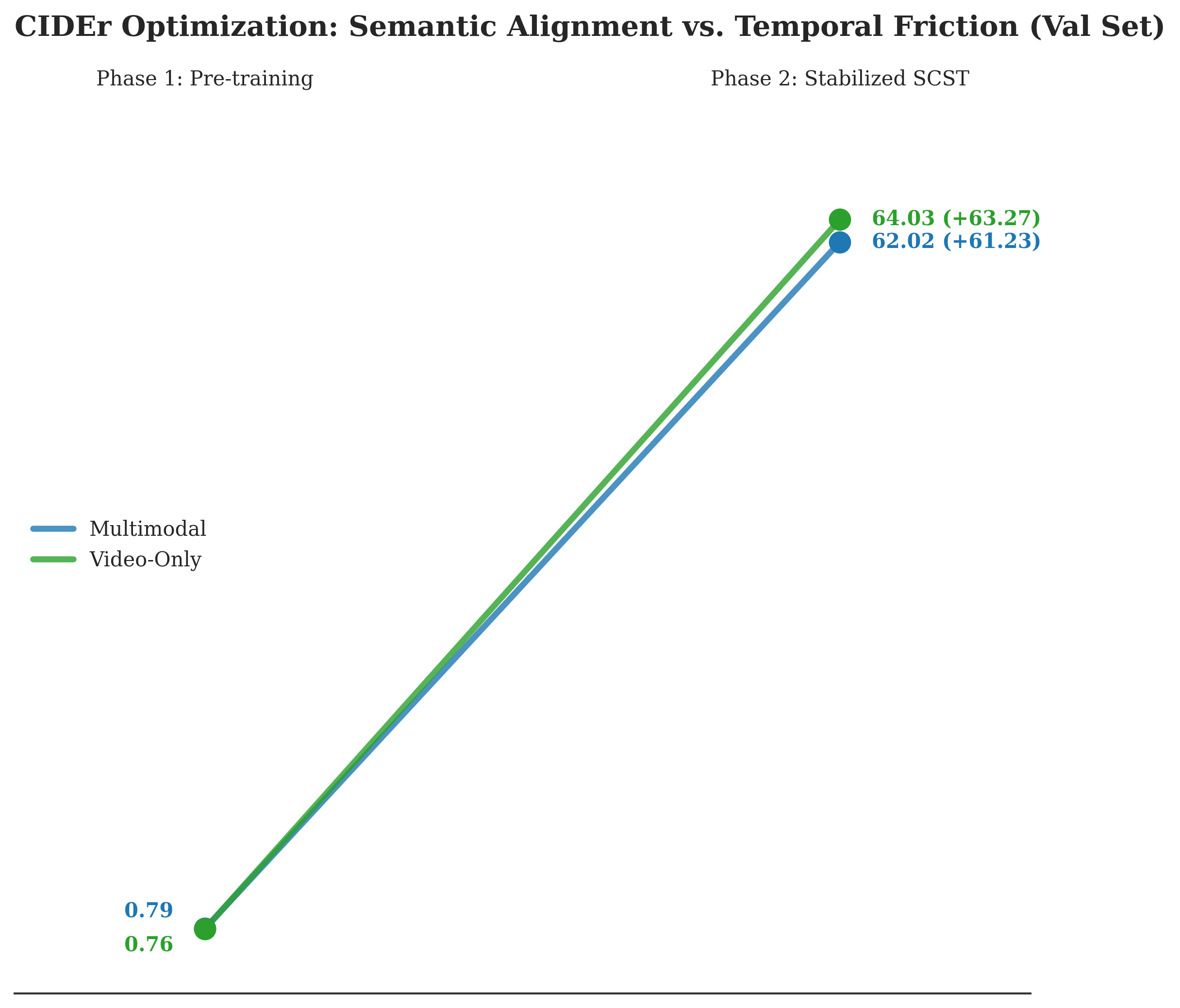}
\caption{Model Performance Progression tracking progression from Phase 1 to Phase 2.}
\label{fig:ablation_slope}
\end{figure}

\begin{figure}[ht]
\centering
\includegraphics[width=0.8\textwidth]{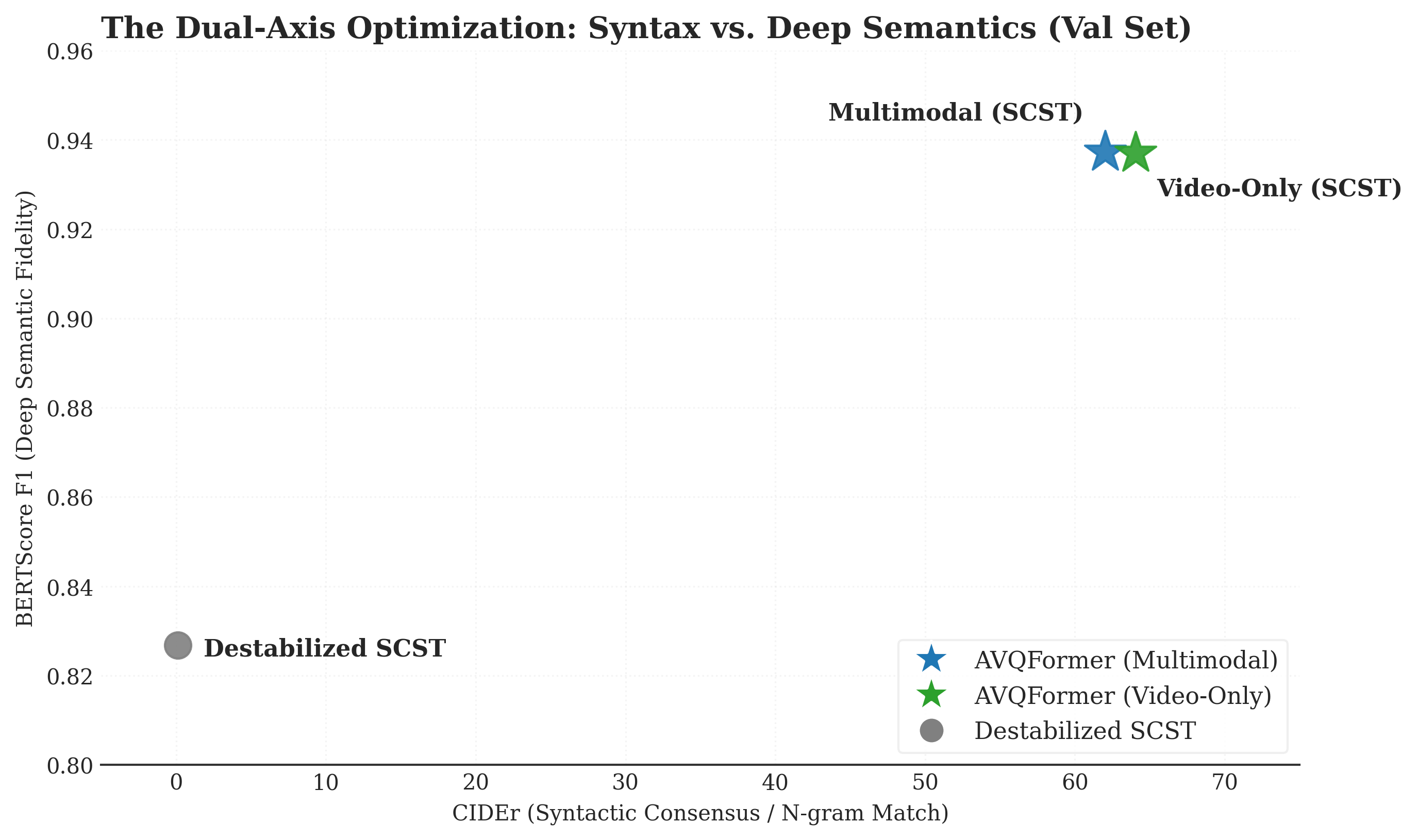}
\caption{Syntax vs. Deep Semantics Optimization on the validation set, showing the divergent strengths of the Video-Only and Multimodal approaches.}
\label{fig:bert_scatter}
\end{figure}

We conducted strict modality and stability ablations on the cleaned MSR-VTT validation set to isolate architectural variables (Table~\ref{tab:ablation}, Figures~\ref{fig:ablation_slope} and \ref{fig:bert_scatter}). Phase 1 utilized standard Cross-Entropy training. The seemingly poor Phase 1 CIDEr score (0.79) is a direct consequence of length mismatch and conversational verbosity of the Phi-2 LLM, which triggers severe length penalties under CIDEr's n-gram matching formulation, despite possessing high semantic fidelity as verified by the robust BERTScore.

Turning off the ASCST clamping and Entropy Regularization caused catastrophic mode collapse, plunging CIDEr to 0.11 and BLEU-4 to 7.48. This empirically demonstrates that our stabilization techniques are critical for bridging the modality gap via RL.

\begin{table}[ht]
\centering
\caption{Architectural Ablations on the Cleaned MSR-VTT Validation Set. Metrics follow the abbreviations defined in Table~\ref{tab:sota_msrvtt}.}
\label{tab:ablation}
\begin{tabular}{@{}lccccc@{}}
\toprule
\textbf{Architecture} & \textbf{C} & \textbf{B@4} & \textbf{M} & \textbf{BS (F1)} & \textbf{BS (Recall)} \\ \midrule
Phase 1 Baseline (Multimodal) & 0.79 & 16.9 & 26.3 & 0.8919 & 0.9233 \\
Phase 2 Destabilized SCST & 0.11 & 7.5 & 21.2 & 0.8268 & 0.9167 \\ \midrule
Phase 2 AVQFormer (Multimodal) & 62.0 & \textbf{46.9} & 31.1 & \textbf{0.9373} & \textbf{0.9376} \\ 
Phase 2 AVQFormer (Video-Only) & \textbf{64.0} & 46.5 & \textbf{31.1} & 0.9371 & 0.9366 \\ \bottomrule
\end{tabular}
\end{table}

\section{Discussion: Investigating the Modality Gap}

\subsection{Phase 1 Prompting: Semantic Alignment of the Baseline}
\label{sec:phase1_prompting}
To explicitly validate our ``semantics-first'' hypothesis, we evaluated the Phase 1 Cross-Entropy models prior to SCST optimization using a rigid conciseness prompt. Deep semantic metrics from this zero-shot baseline completely upend the syntactic narrative of Table~\ref{tab:sota_msrvtt}. In this pre-RL phase, the prompted Multimodal Base model actively outperformed the prompted Video-Only model across semantic metrics, including METEOR (25.3 vs.\ 25.1) as well as BERTScore F1 (0.889 vs.\ 0.888), differing slightly from the unprompted baseline in Table 2. This confirms that the audio pipeline successfully extracts and injects valid, highly accurate semantic concepts into the latent space.

However, the prompted evaluation also revealed a persistent length mismatch: the models generated roughly 87,000 words compared to the 51,200-word ground truth reference (a generation ratio of $\sim 1.69$). Prompt engineering alone is insufficient to override the pre-trained conversational verbosity of the Phi-2 LLM, resulting in heavily penalized n-gram scores across the board (CIDEr $\sim$0.85).

We hypothesize this dichotomy is driven by \textit{temporal alignment friction}. The Multimodal architecture demonstrates strong deep semantic comprehension. However, to survive the rigid, visually-biased syntax required by legacy n-gram metrics like CIDEr, the model must undergo SCST. During SCST, the forced constraint of visual conformity punishes the unaligned audio features, forcing the model into a syntactic compromise. The Video-Only model ultimately achieves the highest CIDEr scores because its unimodal nature allows it to seamlessly adapt to the visually-biased reward function without auditory interference.

\subsection{The Reality of Modality Interference under RL}
The superior performance of the Video-Only model over the Multimodal architecture across syntactic metrics post-SCST sheds light on this performance gap. When humans annotate datasets like MSR-VTT, they exhibit a strong visual bias, detailing what is seen rather than heard. In our Phase 2 validations (Table~\ref{tab:ablation}), adding audio features (CLAP~\cite{Wu2023}) to the AVQFormer slightly improved BERTScore Recall (0.9366 to 0.9376), indicating that accurate, factual acoustic concepts were successfully retained. However, this same injection lowered CIDEr (64.0 to 62.0). 

Because our $K=16$ temporal bottleneck provides the LLM decoder with a pristine, chronologically ordered sequence of frames, the forced concatenation of an unaligned, global audio token introduces modality interference under RL. It is important to note that this friction may be exacerbated by the global-to-local architectural mismatch (fusing a single global audio token against 16 temporally dense visual tokens) and a strong visual bias of the MSR-VTT dataset. The LLM allocates attention to acoustic features that disrupt the rigid, minimal syntax rewarded by SCST (as demonstrated in Table~\ref{tab:friction_examples}).

\begin{table}[ht]
\centering
\caption{Qualitative Examples of Temporal Alignment Friction. The Multimodal Base model extracts auditory context (highlighted in \textit{italics}), but is penalized because the visually-biased ground truths omit these events. \textbf{Note:} These qualitative examples were specifically selected to illustrate instances where the inclusion of audio features negatively impacts the n-gram score by adding non-visual context.}
\label{tab:friction_examples}
\begin{tabular}{@{}>{\raggedright\arraybackslash}p{0.12\textwidth} >{\raggedright\arraybackslash}p{0.28\textwidth} >{\raggedright\arraybackslash}p{0.28\textwidth} >{\raggedright\arraybackslash}p{0.24\textwidth}@{}}
\toprule
\textbf{Video ID} & \textbf{Video-Only (High CIDEr)} & \textbf{Multimodal Base (Low CIDEr)} & \textbf{Selected Ground Truths} \\ \midrule

\textbf{video7624} \newline (Dance) & 
a man and woman are dancing together on a stage in front of an audience\dots & 
a man and a woman are dancing on a stage\dots \textit{while a man narrates what they are doing}\dots & 
(1) a man and a woman are dancing in a stage \newline (2) stage actors dancing on stage \\ \midrule

\textbf{video7050} \newline (Stroller) & 
this is a video of a baby stroller being assembled and disassembled by a man\dots & 
\textit{a woman is talking about a baby stroller} while a stroller is being displayed\dots & 
(1) a person is pushing a stroller \newline (2) a woman demonstrates how to fold up a stroller \\ \midrule

\textbf{video7979} \newline (Boxing) & 
two men are boxing each other in a ring in front of a large crowd of people\dots & 
two men are boxing each other in a ring\dots \textit{while a man narrates what is going on}\dots & 
(1) two men are boxing in the ring \newline (2) two men are engaged in a boxing match \\ \bottomrule
\end{tabular}
\end{table}

\subsection{Architectural Explorations: The Limits of Modality Gating}
During architectural development, we explored a dynamically gated variant of the AVQFormer designed to explicitly route visual and audio features to minimize interference by utilizing a learnable scalar weight to modulate the audio token's attention contribution before fusing it with the visual queries. However, evaluating this Gated Model in the Phase 1 prompted baseline revealed a marked decrease in syntactic consensus (0.79 CIDEr) compared to the Multimodal Base Model (0.85 CIDEr). This early degradation indicates that the gating mechanism itself introduces representational noise and optimization bottlenecks even before Reinforcement Learning is applied. When subjected to the harsh, visually-biased reward landscape of Phase 2 SCST, the Gated Model experienced catastrophic mode collapse due to credit assignment failures. This suggests that the baseline AVQFormer, allowing the LLM's inherent cross-attention to naturally fuse the temporal frames and global audio token, may be superior to hard-coding a restrictive routing mechanism in this context.

\begin{figure}[H]
\centering
\begin{tabular}{p{0.95\linewidth}}
\toprule
\textbf{Video ID: video7285 (Cooking)} \\ \midrule
{\centering \includegraphics[width=0.45\textwidth]{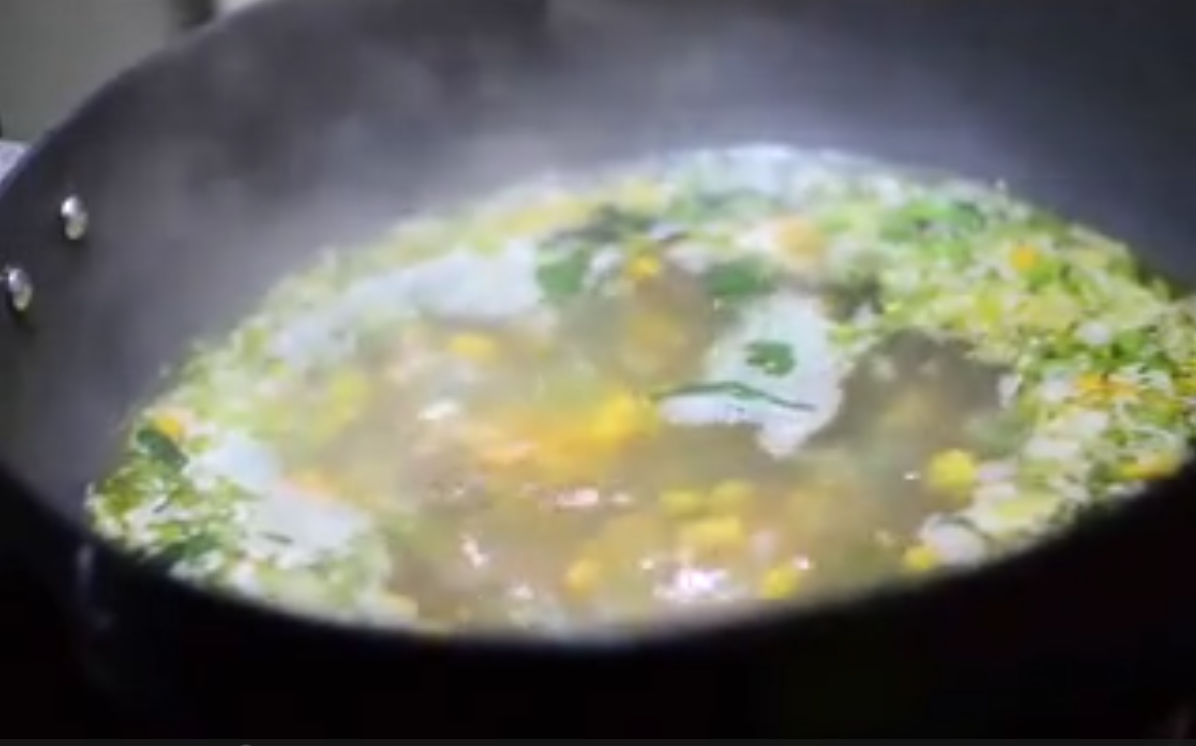} \par}
\textbf{Proposed AVQFormer (Video-Only):} a man is cooking a soup in a pot on the stove \\
\textbf{Proposed AVQFormer (Multimodal Base):} a person is cooking a soup in a pot on the stove \\
\textbf{Ground Truth Sample:} \newline
(1) a man cooks various ingredients in a pot on the stove \newline
(2) in a kitchen someone is cooking a soup in the black pan \newline
(3) on a stove top a pot is boiling vegetables while a chef adds ingredients \\ \bottomrule
\end{tabular}
\caption{Qualitative comparison of generated captions for a cooking video (video7285). The Video-Only model generates specific visual nouns (e.g., ``man'') that align directly with the ground truth, whereas the Multimodal model generates broader semantic terms (e.g., ``person'').}
\label{fig:qualitative_cooking}
\end{figure}

\begin{figure}[H]
\centering
\begin{tabular}{p{0.95\linewidth}}
\toprule
\textbf{Video ID: video7280 (Fish)} \\ \midrule
{\centering \includegraphics[width=0.45\textwidth]{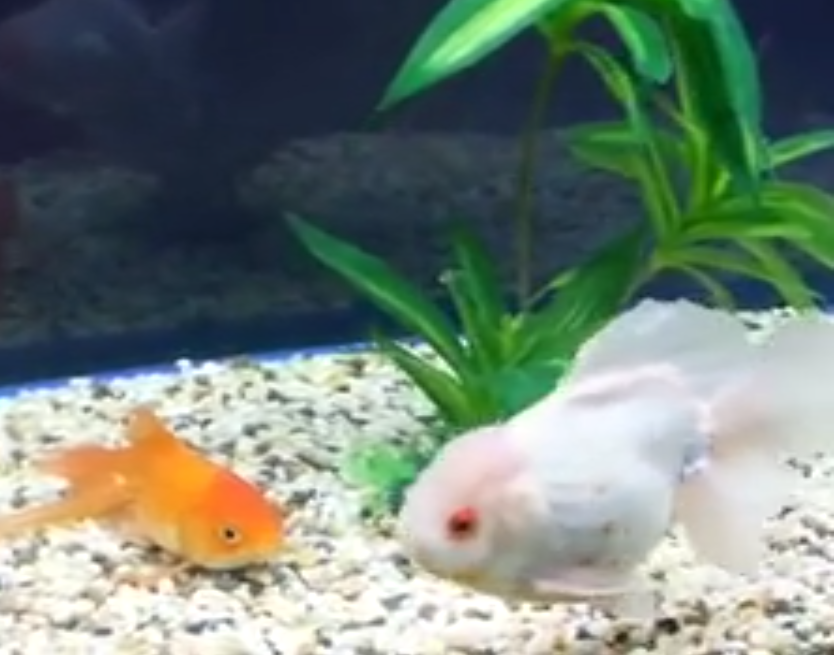} \par}
\textbf{Proposed AVQFormer (Video-Only):} a group of fish are swimming in a tank \\
\textbf{Proposed AVQFormer (Multimodal Base):} a fish is swimming in the water \\
\textbf{Ground Truth Sample:} \newline
(1) different gold fish are swimming around in a tank opening and closing their mouths \newline
(2) golden fish in glass tank \newline
(3) a fish tank with a fish inside of it swimming past orange flowers \\ \bottomrule
\end{tabular}
\caption{Qualitative comparison of generated captions for an aquarium video (video7280). The Multimodal model omits specific environmental nouns (e.g., ``tank'') present in the ground truth and Video-Only outputs, utilizing more generalized vocabulary (e.g., ``water'').}
\label{fig:qualitative_fish}
\end{figure}

As demonstrated by the specific visual nouns generated without audio (Figure~\ref{fig:qualitative_cooking}) and the generic semantic diffusion observed when audio is added (Figure~\ref{fig:qualitative_fish}), the introduction of global audio mathematically penalizes the model under exact n-gram matching. Consequently, we determine that for highly distilled visual architectures, unimodal visual input currently yields the highest alignment with visually-biased captioning benchmarks.

\subsection{Limitations and Future Work}
While our approach yields highly competitive performance, certain limitations remain. Our current evaluation is limited to a single primary benchmark (MSR-VTT); future work must explore cross-dataset generalization (e.g., VATEX, ActivityNet Captions) to fully validate the architecture's robustness against dataset-specific overfitting. Additionally, because the architecture relies entirely on the generalized knowledge of frozen CLIP and CLAP encoders, any inherent representational biases within those foundation models are passed directly to the language decoder. Furthermore, while utilizing a single global audio token from CLAP provides computational efficiency, this intentional bottleneck heavily restricts the model's ability to map chronological grammar to isolated acoustic events, which likely explains the catastrophic failure of the Audio-Only ablation (0.14 CIDEr). Finally, qualitative failure modes indicate that while the model excels at broad semantic understanding, it can occasionally miss highly fine-grained visual interactions or decouple auditory context from visual events. Future work will explore scaling to larger, more diverse datasets and refined temporal tokenization strategies to capture these micro-interactions.

\clearpage
\section{Conclusion}
We introduce the AVQFormer, demonstrating that a highly efficient multimodal architecture can achieve strong deep semantic comprehension. Through zero-shot prompted evaluations, we observed evidence that global audio tokens successfully inject rich, non-destructive semantic context into the latent space. However, by deploying Adaptive SCST, we exposed the potential for temporal alignment friction inherent in legacy evaluation metrics. Because datasets like MSR-VTT demand rigid, visually-biased syntax, the RL pipeline actively penalizes auditory intelligence. This forces a paradigm where unimodal visual inputs (achieving a highly competitive 63.0 CIDEr among lightweight models) artificially outscore highly capable multimodal agents on n-gram benchmarks. Ultimately, this work establishes a highly compute-efficient standard for generative video research while demonstrating that metrics like CIDEr, when applied to datasets such as MSR-VTT, exhibit a strong bias toward visually grounded syntax, challenging the field to look beyond these measures when evaluating true multimodal understanding.

\section*{Data and Code Availability}
To support independent reproducibility and further research within the multimodal video understanding community, all training code, evaluation scripts, and pretrained model weights will be made publicly available upon publication.

\end{document}